\pdfoutput=1

\documentclass[11pt]{article}

\usepackage[]{emnlp2021}

\usepackage{times}
\usepackage[T1]{fontenc}
\usepackage[utf8]{inputenc}
\usepackage{microtype}
\usepackage{amsmath}
\usepackage{xcolor}
\usepackage{ragged2e}
\usepackage{bm}
\usepackage{latexsym}
\usepackage{subfigure}
\usepackage{graphicx}
\usepackage{float}
\usepackage{longtable}
\usepackage{booktabs} 
\usepackage{multirow}
\usepackage{bbding} 

\usepackage{amssymb}
\usepackage[ruled,vlined]{algorithm2e}
\usepackage{color, soul}

\title{PTR: Prompt Tuning with Rules for Text Classification}

\author{Xu Han, Weilin Zhao, Ning Ding, \textbf{Zhiyuan Liu}\thanks{\quad Corresponding author: Z.Liu(liuzy@tsinghua.edu.cn)}\hspace{0.5em}, \textbf{Maosong Sun}\\
 State Key Lab on Intelligent Technology and Systems,\\
Institute for Artificial Intelligence, \\
Department of Computer Science and Technology, Tsinghua University, Beijing, China \\
\texttt{\{hanxu17,zwl19,dingn18\}@mails.tsinghua.edu.cn}\\
\texttt{\{liuzy,sms\}@tsinghua.edu.cn}}

\begin{document}
\maketitle
\begin{abstract}

Fine-tuned pre-trained language models (PLMs) have achieved awesome performance on almost all NLP tasks. 
By using additional prompts to fine-tune PLMs, we can further stimulate the rich knowledge distributed in PLMs to better serve downstream tasks. 
Prompt tuning has achieved promising results on some few-class classification tasks such as sentiment classification and natural language inference. 
However, manually designing lots of language prompts is cumbersome and fallible. 
For those auto-generated prompts, it is also expensive and time-consuming to verify their effectiveness in non-few-shot scenarios. 
Hence, it is still challenging for prompt tuning to address many-class classification tasks. 
To this end, we propose prompt tuning with rules (PTR) for many-class text classification and apply logic rules to construct prompts with several sub-prompts. 
In this way, PTR is able to encode prior knowledge of each class into prompt tuning. 
We conduct experiments on relation classification, a typical and complicated many-class classification task, and the results show that PTR can significantly and consistently outperform existing state-of-the-art baselines. 
This indicates that PTR is a promising approach to take advantage of both human prior knowledge and PLMs for those complicated classification tasks.
\end{abstract}

\section{Introduction}

Recently, pre-trained language models (PLMs) such as GPT~\cite{radfordimproving}, BERT~\cite{devlin2019bert}, RoBERTa~\cite{liu2019roberta}, and T5~\cite{raffel2020exploring} have emerged as a powerful instrument for language understanding and generation. By applying self-supervised learning, PLMs can capture a wealth of linguistic~\cite{jawahar2019does}, semantic~\cite{yenicelik2020does}, syntactic~\cite{hewitt2019structural}, and world knowledge~\cite{petroni2019language} from large-scale corpora, which is the key to the success of PLMs. By fine-tuning these PLMs with additional task-specific data, rich knowledge distributed in PLMs can be adapted to various downstream tasks. In the past few years, PLM fine-tuning has shown awesome performance on almost all important NLP tasks. It is now a consensus of the NLP community to fine-tune PLMs for specific tasks instead of learning models from scratch~\cite{qiu2020pre}. 

\begin{figure*}[t]
    \centering
    \includegraphics[width = 1.0\linewidth]{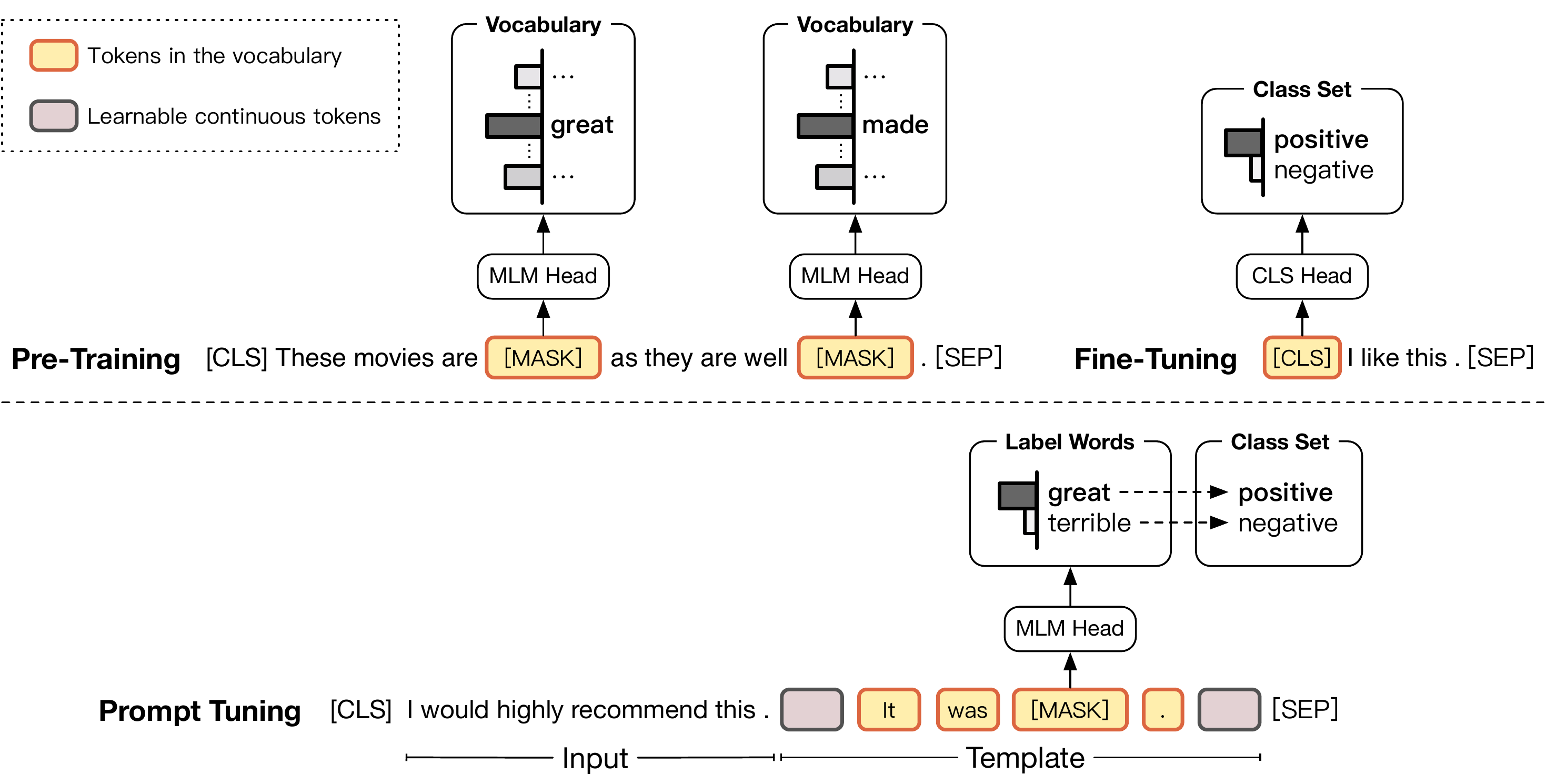}
    \caption{The example of pre-training, fine-tuning, and prompt tuning. The blue rectangles in the figure are special prompt tokens, whose parameters are randomly initialized and learnable during prompt tuning.}
    \label{fig:example}
\end{figure*}

Despite the success of fine-tuning PLMs, some recent studies find one of its critical challenges is the significant gap of objective forms in pre-training and fine-tuning, which restricts taking full advantage of knowledge in PLMs. Take Figure~\ref{fig:example} for example, pre-training is usually formalized as a cloze-style task to predict target words (e.g. sequential language models and masked language models), yet downstream tasks in fine-tuning may exhibit different objective forms such as classification, generation, and sequence labeling. This gap will hinder the transfer and adaptation of knowledge in PLMs to downstream tasks.

To alleviate this issue, prompt tuning has been proposed to bridge the gap of objective forms in pre-training and fine-tuning~\cite{brown2020language,schick2020automatically}. As shown in Figure~\ref{fig:example}, a typical prompt consists of a template (e.g. ``<$S_1$> It was $\texttt{[MASK]}$.'') and a set of label words (e.g. ``great'' and ``terrible''). The set of label words serves as the candidate set for predicting $\texttt{[MASK]}$. By fusing the original input with the prompt template for predicting $\texttt{[MASK]}$ and then mapping predicted words to corresponding labels, prompt tuning can convert a binary sentiment classification task into a cloze-style task. In other words, by checking that PLMs predict ``great'' or ``terrible'' at the masked position, we can determine whether <$S_1$> is positive or negative. Besides sentiment classification, existing prompt tuning methods have also achieved promising results on some other few-class classification tasks such as natural language inference~\cite{schick2020exploiting,liu2021gpt}. For instance, by using ``<$S_1$> . \texttt{[MASK]} , <$S_2$>'' as the prompt template, and $\{\text{``yes''}, \text{``maybe''}, \text{``no''}\}$ as the set of label words, we can also easily use PLMs to predict the discourse relation (entailment, neutral, or contradiction) between <$S_1$> and <$S_2$>.

However, for those tasks with many classes, it becomes challenging to manually find appropriate templates and suitable label words to distinguish different classes. For example, relation classification, a typical many-class classification task, requires models to predict semantic relations between two marked entities in the text. Given the relation ``person:parent'' and the relation ``organization:parent'', it is hard to pick label words to distinguish them. A straightforward solution is automatically generating prompts: \citet{schick2020automatically,schick2020exploiting} automatically identify label words from the vocabulary for human-picked templates; \citet{shin2020eliciting} utilize gradient-guided search to automatically generate both templates and label words; \citet{gao2020making} propose to generate all prompt candidates and apply development sets to find the most effective ones. Although auto-generated prompts can avoid intensive labor, it is not surprising that most auto-generated prompts cannot achieve comparable performance to human-picked ones. Meanwhile, auto-generated prompts require extra computation costs for generation and verification. The expensive extra computation costs make auto-generated prompts more suitable for few-shot learning, rather than those normal learning settings with many instances and classes.

In this paper, we propose prompt tuning with rules (PTR) for many-class classification tasks. For a many-class classification task, we manually design essential sub-prompts and apply logic rules to compose sub-prompts into final task-specific prompts. Compared with existing prompt tuning methods, PTR has two advantages:

(1) \textbf{Prior Knowledge Encoding}. 
PTR can apply logic rules to encode prior knowledge about tasks and classes into prompt tuning. We also take relation classification as an example, the predicted results are usually correlated to both relational semantics of sentences and types of entities. We can build the prompt for the relation ``person:parent'' and the relation``organization:parent'' with two sub-prompts, one sub-prompt is to determine whether marked entities are human beings or organizations, and the other sub-prompt is to determine whether sentences express the semantics of the parent-child relationship. 

(2) \textbf{Efficient Prompt Design}. 
Compared with manually designing templates and individual label words for all classes, it is easier to design several simple sub-prompts and then combine these sub-prompts to form task-specific prompts according to logic rules. Moreover, compared with automatically generating prompts based on machine learning models, using logic rules to generate prompts is more efficient and interpretable.


To verify the effectiveness of PTR, we conduct experiments on relation classification, a typical and complicated many-class classification task. Four popular benchmarks of relation classification are used for our experiments, including TACRED~\cite{zhang2017position}, TACREV~\cite{alt2020tacred}, ReTACRED~\cite{stoica2021re}, and SemEval 2010 Task 8~\cite{hendrickx2010semeval}. Sufficient experimental results show that PTR can significantly and consistently outperform existing state-of-the-art baselines.

\begin{figure*}[t]
    \centering
    \includegraphics[width = 1.0\linewidth]{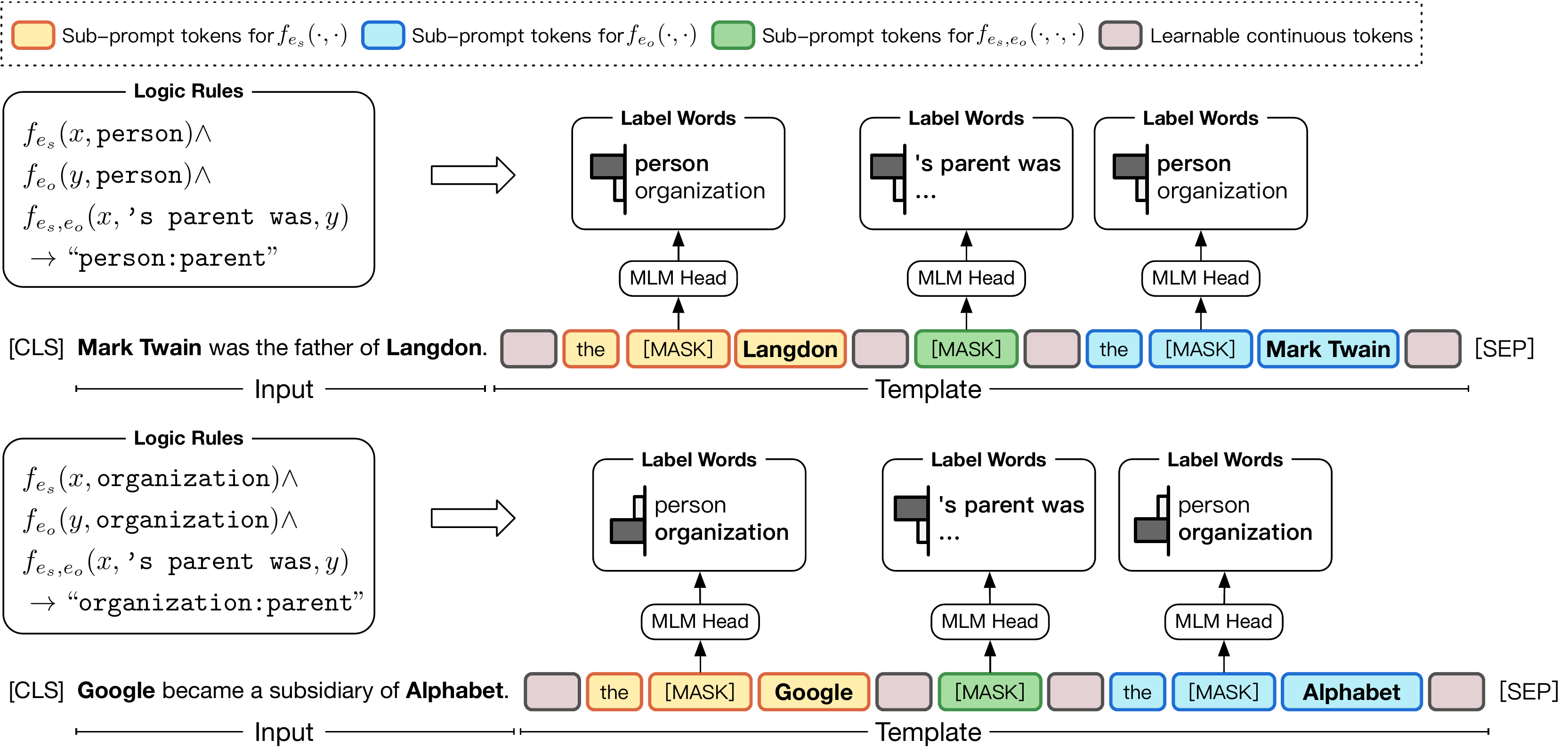}
    \caption{Illustration of PTR. By composing the sub-prompts of the conditional functions
     $f_{e_s}(\cdot,\cdot)$, $f_{e_o}(\cdot,\cdot)$ and $f_{e_s,e_o}(\cdot,\cdot,\cdot)$, 
     we could easily get an effective prompt to distinguish ``person:parent'' and ``organization:parent''.
     }
    \label{fig:intro}
\end{figure*}

\section{Preliminaries}
\label{sec:setup}

Before introducing PTR, we first give some essential preliminaries, especially some details about prompt tuning for classification tasks. Formally, a classification task can be denoted as $\mathcal{T} = \{\mathcal{X}, \mathcal{Y}\}$, where $\mathcal{X}$ is the instance set and $\mathcal{Y}$ is the class set. For each instance $x \in \mathcal{X}$, it consists of several tokens $x = \{w_1, w_2, \ldots, w_{|x|}\}$, and is annotated with a label $y_x \in \mathcal{Y}$. 

\subsection{Vanilla Fine-tuning for PLMs} 

Given a PLM $\mathcal{M}$, vanilla fine-tuning first converts the instance $x=\{w_1, w_2, \ldots, w_{|x|} \}$ to the input sequence $\{\texttt{[CLS]}, w_1, w_2, \ldots, w_{|x|}, \texttt{[SEP]}\}$, and then uses $\mathcal{M}$ to encode all tokens of the input sequence into corresponding vectors $\{\mathbf{h}_{\texttt{[CLS]}}, \mathbf{h}_{w_1}, \mathbf{h}_{w_2}, \ldots, \mathbf{h}_{w_{|x|}}, \mathbf{h}_{\texttt{[SEP]}}\}$. For a downstream classification task, a task-specific head is used to compute the probability distribution over the class set $\mathcal{Y}$ with the softmax function $p(\cdot|x) = \texttt{Softmax}(\mathbf{W}\mathbf{h}_{\texttt{[CLS]}}+\mathbf{b})$, where $\mathbf{h}_{\texttt{[CLS]}}$ is the hidden vector of $\texttt{[CLS]}$, $\mathbf{b}$ is a learnable bias vector, and $\mathbf{W}$ is a learnable matrix randomly initialized before fine-tuning. The parameters of $\mathcal{M}$, $\mathbf{b}$, and $\mathbf{W}$ are tuned to maximize $\frac{1}{|\mathcal{X}|} \sum_{x\in \mathcal{X}} \log p(y_x|x)$.

\subsection{Prompt Tuning for PLMs}

To apply cloze-style tasks to tune PLMs, prompt tuning has been proposed. Formally, a prompt consists of a template $T(\cdot)$ and a set of label words $\mathcal{V}$. For each instance $x$, the template is first used to map $x$ to the prompt input $x_\text{prompt} = T(x)$. The template defines where each token of $x$ is placed and whether to add any additional tokens. Besides retaining the original tokens in $x$, at least one $\texttt{[MASK]}$ is placed into $x_\text{prompt}$ for $\mathcal{M}$ to fill label words. For instance, for a binary sentiment classification task, we set a template $T(\cdot) = \text{``} \cdot \text{It was} \texttt{[MASK]} \text{.''}$, and map $x$ to $x_\text{prompt} = \text{``} x \text{ It was} \texttt{[MASK]} \text{.''}$. 

By input $x_\text{prompt}$ into $\mathcal{M}$, we compute the hidden vector $\mathbf{h}_{\texttt{[MASK]}}$ of $\texttt{[MASK]}$. Given $v \in \mathcal{V}$, we produce the probability that the token $v$ can fill the masked position $p(\texttt{[MASK]}=v|x_\text{prompt}) = \frac{\exp(\mathbf{v}\cdot \mathbf{h}_{\texttt{[MASK]}})}{\sum_{\tilde{v}\in\mathcal{V}}\exp(\mathbf{\tilde{v}}\cdot \mathbf{h}_{\texttt{[MASK]}})}$, where $\mathbf{v}$ is the embedding of the token $v$ in the PLM $\mathcal{M}$.

There also exists an injective mapping function $\phi: \mathcal{Y} \rightarrow \mathcal{V}$ that bridges the set of classes and the set of label words. In some papers, the function $\phi$ is named ``verbalizer''. With the verbalizer $\phi$, we can formalize the probability distribution over $\mathcal{Y}$ with the probability distribution over $\mathcal{V}$ at the masked position, i.e. $p(y|x) = p([\texttt{MASK}] = \phi(y) | x_\text{prompt})$. Here, we also take a binary sentiment classification task as an example. We map the positive sentiment to ``great'' and the negative sentiment to ``terrible''. According to $\mathcal{M}$ fills the masked position of $x_\text{prompt}$ with ``great'' or ``terrible'', we can know the instance $x$ is positive or negative. 

For prompt tuning, with the template $T(\cdot)$, the set of label words $\mathcal{V}$, and the verbalizer $\phi$, the learning objective is to maximize $\frac{1}{|\mathcal{X}|} \sum_{x\in \mathcal{X}} \log p([\texttt{MASK}] = \phi(y_x)|T(x))$. In Figure~\ref{fig:example}, we show pre-training, fine-tuning, and prompt tuning, which can indicate the connections and differences among them clearly.

\section{Prompting Tuning with Rules (PTR)}
\label{sec:prompt}

As we mentioned before, prompt tuning is well developed for few-class classification tasks such as sentiment classification and natural language inference. Considering the difficulty of designing prompts for many-class classification tasks, we propose PTR to incorporate logic rules to compose task-specific prompts with several simple sub-prompts. In this section, we will introduce the details of PTR, and select relation classification as an example to illustrate how to adapt PTR for many-class classification tasks.

\subsection{Overall Framework of PTR}

PTR is driven by basic human logic inferences. For example, in relation classification, if we wonder whether two marked entities in a sentence have the relation ``person:parent'', with prior knowledge, we would check whether the sentence and two marked entities meet certain conditions: (1) \textit{The two marked entities are human beings}; (2) \textit{the sentence indicates the parental semantics between the two marked entities}. 

Inspired by this, for any text classification task $\mathcal{T} = \{\mathcal{X}, \mathcal{Y}\}$, we design a conditional function set $\mathcal{F}$. Each conditional function $f \in \mathcal{F}$ determines whether the function input meets certain conditions. For instance, the conditional function $f(x, \texttt{person})$ can determine whether the input $x$ is a person, the conditional function $f(x, \texttt{'s parent was}, y)$ can determine whether $y$ is the parent of $x$. Intuitively, these conditional functions are essentially the predicates of first-order logic.

For each conditional function $f \in \mathcal{F}$, PTR set a template $T_f(\cdot)$ and a set of label words $\mathcal{V}_f$ to build its sub-prompt. According to the semantics of $\mathcal{Y}$, we can use logic rules to transform the classification task into the calculation of a series of conditional functions. As shown in Figure~\ref{fig:intro}, for relation classification, determining whether the entities $x$ and $y$ have the relation ``person:parent'' can be formalized as 
\begin{equation}
\small
\begin{aligned}
&f_{e_s}(x, \texttt{person}) \wedge f_{e_s,e_o}(x,\texttt{'s parent was},y) \\ 
&\wedge f_{e_o}(y, \texttt{person}) \rightarrow \text{``person:parent''},\nonumber
\end{aligned}
\end{equation}
where $f_{e_s}(\cdot,\cdot)$ is the conditional function to determine the subject entity types, $f_{e_o}(\cdot,\cdot)$ is the conditional function to determine the object entity types, and $f_{e_s,e_o}(\cdot,\cdot,\cdot)$ is the conditional function to determine the semantic connection between entities. Similarly, determining the relation ``organization:parent'' can be formalized as 
\begin{equation}
\small
\begin{aligned}
&f_{e_s}(x, \texttt{organization}) \\
&\wedge f_{e_s,e_o}(x,\texttt{'s parent was},y) \\ 
&\wedge f_{e_o}(y, \texttt{organization}) \\
&\rightarrow \text{``organization:parent''}.\nonumber
\end{aligned}
\end{equation}
Based on the above-mentioned logic rules and conditional functions, we can compose the sub-prompts of rule-related conditional functions to handle the task $\mathcal{T}$. Next, we will introduce how to design sub-prompts for conditional functions, as well as how to aggregate sub-prompts for tasks.

\subsection{Sub-Prompts for Conditional Functions}

For each conditional function $f\in\mathcal{F}$, we manually design a sub-prompt for it. Similar to the conventional prompt setting, a sub-prompt also consists of a template and a set of label words. 

The basic conditional function is a unary function. For binary sentiment classification, unary functions can determine whether a sentence is positive or negative; For entity typing, unary functions can indicate entity types; For topic labeling, unary functions can be used to predict article topics. 

Here we still take relation classification as an example. Given a sentence $x = \{\ldots e_s \ldots e_o \ldots\}$, where $e_s$ and $e_o$ are the subject entity and the object entity respectively, we set $f_{e_s}(\cdot,\{\text{person}|\text{organization}|\ldots\})$ to determine the type of $e_s$. The sub-prompt template and label word set of $f_{e_s}$ can be formalized as
\begin{equation}
\small
\label{eq:sub-prompt-e1}
\begin{aligned}
& T_{f_{e_s}}(x) = \text{``} x \text{ the} \texttt{[MASK]} e_s \text{''},\\
& \mathcal{V}_{f_{e_s}} = \{ \text{``person''}, \text{``organization''}, \ldots\}.
\end{aligned}
\end{equation}
For the object entity $e_o$, the sub-prompt template and label word set of $f_{e_o}$ are similar to Eq.~(\ref{eq:sub-prompt-e1}).

Another important type of conditional functions is a binary function. For natural language inference, binary functions can determine the relation between two sentences is entailment, neutral, or contradiction; For relation classification, such functions can be used to determine the complex connections between two entities. If we set $f_{e_s,e_o}(\cdot,\{\text{'s parent was}|\text{was born in}|\ldots \},\cdot)$, the sub-prompt template and label word set of $f_{e_s,e_o}$ can be formalized as
\begin{equation}
\small
\label{eq:sub-prompt-e1e2}
\begin{aligned}
& T_{f_{e_s,e_o}}(x) = \text{ `` } x \text{  }  e_s \texttt{[MASK]} e_o \text{''},\\
& \mathcal{V}_{f_{e_s,e_o}} = \{ \text{``'s parent was''}, \text{``was born in''}, \ldots\}.
\end{aligned}
\end{equation}

Normally, unary and binary functions are sufficient to compose logic rules for classification tasks. For classification tasks with complex semantics, these design methods of unary and binary functions can be extended to multi-variable functions, which can provide more powerful sub-prompts.

\subsection{Composing Sub-Prompts for Tasks}

Since we transform a classification task into the calculation of a series of conditional functions, we need to compose sub-prompts of various conditional functions into a complete task-specific prompt. In this paper, we apply a simple strategy: we use logic rules with conjunctive normal form and then directly concatenate the sub-prompts of all rule-related functions. 

For instance, in Figure~\ref{fig:intro}, by aggregating the sub-prompts in Eq.~(\ref{eq:sub-prompt-e1}) and Eq.~(\ref{eq:sub-prompt-e1e2}), the complete prompt template is as follows,
\begin{equation}
\small
\begin{aligned}
T(x) &= [T_{f_{e_s}}(x) ; T_{f_{e_s,e_o}}(x); T_{f_{e_o}}(x)] = \\
\text{ `` } &x \text{ the } \texttt{[MASK]}_1~ e_s  \texttt{[MASK]}_2  \text{ the } \texttt{[MASK]}_3~ e_o \text{ '' }, 
\end{aligned}
\end{equation}
where $[\cdot;\cdot;\cdot]$ is the aggregation function of sub-prompts. And the aggregated sets of label words are given as
\begin{equation}
\small
\begin{aligned}
\mathcal{V}_{\texttt{[MASK]}_1} &= \{\text{``person''}, \text{``organization''},\ldots\},\\
\mathcal{V}_{\texttt{[MASK]}_2} &= \{\text{``'s parent was''}, \text{``was born in''},\ldots\},\\
\mathcal{V}_{\texttt{[MASK]}_3} &= \{\text{``person''}, \text{``organization''},\ldots\}.
\end{aligned}
\end{equation}
As shown in Figure~\ref{fig:intro}, we can also add some learnable tokens, whose parameters are randomly initialized, to make the template more effective.

As the aggregated template may contain multiple $\texttt{[MASK]}$ tokens, we must consider all masked positions to make predictions, i.e.,
\begin{equation}
\label{eq:loss}
\small
p(y|x) = \prod_{j=1}^{n} p(\texttt{[MASK]}_j = \phi_j(y)| T(x)),
\end{equation}
where $n$ is the number of masked positions in $T(x)$, and $\phi_j(y)$ is to map the class $y$ to the set of label words $\mathcal{V}_{\texttt{[MASK]}_j}$ for the $j$-th masked position $\texttt{[MASK]}_j$. Eq.~(\ref{eq:loss}) can be used to tune PLMs and classify classes. The final learning objective of PTR is to maximize 
\begin{equation}
\small
\frac{1}{|\mathcal{X}|} \sum_{x\in \mathcal{X}} \log \prod_{j=1}^{n} p\big(\texttt{[MASK]}_j = \phi_j(y)| T(x)\big) .
\end{equation}

\begin{table}[t]
\small
\center
\scalebox{1.0}{
\begin{tabular}{l|r|r|r|r}
\toprule
Dataset          & \#train & \#dev & \#test & \#rel \\
\midrule
SemEval     &6,507&1,493&2,717&19\\
TACRED      &68,124&22,631&15,509&42\\
TACREV      &68,124&22,631&15,509&42\\
ReTACRED   &58,465&19,584&13,418&40\\
\bottomrule
\end{tabular}}
\caption{Statistics of different datasets.} 
\label{tab:tacreddata} 
\end{table}

\section{Experiments}

We conduct experiments on relation classification to show the effectiveness of PTR on many-class classification tasks. 

\begin{table*}[t]
\center
\small
\scalebox{0.95}{
\begin{tabular}{l|c|c|c}
\toprule
Class Label        & $\texttt{[MASK]}_1$ & $\texttt{[MASK]}_2$ & $\texttt{[MASK]}_3$ \\
\midrule
per:country\_of\_birth & person & was born in & country \\
per:stateorprovince\_of\_birth & person & was born in & state \\
per:city\_of\_birth & person & was born in & city \\
per:employee\_of & person & 's employee was & organization \\
per:parents & person & 's parent was & person \\
per:age & person & 's age was & number \\
org:founded\_by & organization & was founded by & person \\
org:country\_of\_headquarters & organization & was located in & country \\
org:stateorprovince\_of\_headquarters & organization & was located in & state \\
org:city\_of\_headquarters & organization & was located in & city \\
org:number\_of\_employees/members & organization & 's employer has & number \\
org:members & organization & 's member was & organization \\
org:parents & organization & 's parent was & organization \\
no\_relation & entity & is irrelevant to & entity \\
\bottomrule
\end{tabular}}
\centering
\caption{For some relations of the datasets TACRED, TACREV, and ReTACRED, the relation-specific label words for the template: $\langle S1 \rangle~The \texttt{[MASK]}_1~\langle E1\rangle~\texttt{[MASK]}_2 ~the~ \texttt{[MASK]}_3~ \langle E2 \rangle$.}
\label{tab:temp_1}
\end{table*}

\begin{table*}[t]
\center
\small
\scalebox{0.95}{
\begin{tabular}{l|c|c|c}
\toprule
Class Label        & $\texttt{[MASK]}_1$ & $\texttt{[MASK]}_2$ & $\texttt{[MASK]}_3$ \\
\midrule
Member-Collection(e1,e2) & member & related & collection \\
Entity-Origin(e1,e2) & entity & related & origin \\
Cause-Effect(e1,e2) & cause & related & effect \\
Component-Whole(e1,e2) & component & related & whole \\
Product-Producer(e1,e2) & product & related & producer \\
Instrument-Agency(e1,e2) & instrument &  related & agency \\
Entity-Destination(e1,e2) & entity & related & destination \\
Content-Container(e1,e2) & content & related & container \\
Message-Topic(e1,e2) & message & related & topic \\
Other & mention & irrelevant & mention \\
\bottomrule
\end{tabular}}
\centering
\caption{For some relations of the dataset SemEval, the relation-specific label words for the template: $\langle S1 \rangle~The \texttt{[MASK]}_1~\langle E1\rangle~ was \texttt{[MASK]}_2~ to~the~ \texttt{[MASK]}_3~ \langle E2 \rangle$.}
\label{tab:temp_2}
\end{table*}

\subsection{Datasets and Experimental Settings}

We carry out our experiments on four relaction classification datasets: TACRED~\cite{zhang2017position}, TACREV~\cite{alt2020tacred}, ReTACRED~\cite{stoica2021re}, and SemEval 2010 Task 8 (SemEval)~\cite{hendrickx2010semeval}. 

\textbf{TACRED:} one of the largest and most widely used datasets for relation classification. It 
is obtained via crowd-sourcing and contains $42$ relation types (including ``no\_relation''). 

\textbf{TACREV:} one dataset built based on the original TACRED dataset. They find out and correct the errors in the original development set and test set of TACRED, while the training set is left intact. 

\textbf{ReTACRED:} another version of TACRED dataset. They address some shortcomings of the original TACRED dataset, refactor its training set, development set, and test set. ReTACRED also modifies a few relation types.

\textbf{SemEval:} a traditional dataset in relation classification 
covering $9$ relations with two directions and one special relation ``Other''.

More details of these datasets are shown in Table~\ref{tab:tacreddata}. For all the above datasets, we use $F_1$ scores as the main metric for evaluation.

\begin{table*}[t]
\center
\small
\scalebox{1.0}{
\begin{tabular}{l|c|c|c|c|c}
\toprule
Model         & Extra Data & TACRED & TACREV & ReTACRED & SEMEVAL \\
\midrule
\multicolumn{6}{c}{Learning models from scratch}               \\
\midrule
\textsc{PA-LSTM}~\cite{zhang2017position}      & w/o & 65.1   & 73.3   & 79.4      & 84.8    \\
\textsc{C-GCN}~\cite{zhang2018graph}         & w/o & 66.3   & 74.6   & 80.3      & -       \\
\midrule
\multicolumn{6}{c}{Fine-tuning pre-trained models}               \\
\midrule
\textsc{RoBERTa\_large}~\cite{liu2019roberta} & w/o & 68.7   & 76.0   & 84.9      & 87.6    \\
\textsc{KnowBERT}~\cite{peters2019knowledge}      & w/    & 71.5   & 79.3   & -         & 89.1    \\
\textsc{MTB}~\cite{baldini-soares-etal-2019-matching}          &  w/ & 70.1   & -      & -         & 89.5    \\
\textsc{SpanBERT}~\cite{joshi2020spanbert}       & w/ & 70.8   & 78.0   & 85.3      & -       \\
\textsc{LUKE}~\cite{yamada2020luke}           & w/  & \underline{72.7}   & 80.6   & 90.3      & -       \\
\midrule
\multicolumn{6}{c}{Prompt tuning pre-trained models}               \\
\midrule
\textsc{PTR}        & w/o & 72.4   & \underline{81.4}   & \underline{90.9}      & \textbf{\underline{89.9}}      \\
\textsc{PTR (Reversed)}   & w/o  & \textbf{75.9}   & \textbf{83.9}   & \textbf{91.9}      & -      \\
\bottomrule
\end{tabular}}
\caption{$F_1$ scores (\%) on the TACRED, TACREV, ReTACRED and SEMEVAL. For \textsc{RoBERTa\_large}, we report the results of the vanilla version~\textsc{RoBERTa\_large} without adding any entity markers. More adaptions of \textsc{RoBERTa\_large} for relation classification will be reported in the next tables. In the ``Extra Data'' column, ``w/o'' means that no additional data is used for pre-training and fine-tuning, yet ``w/'' means that extra data or knowledge bases are used for data augmentation. The best results are bold, and the best results under normal settings (without reversing relations) are underlined.}
\label{tab:all}
\end{table*}

\subsection{Implementation Details}
\label{sec:imp}

Our model is implemented based on the toolkit Transformers~\cite{wolf2020transformers} and OpenNRE~\cite{han-etal-2019-opennre}. Most hyper-parameters are set following previous works: our model is optimized with Adam using the learning rate of $3e-5$ on \textsc{RoBERTa\_large}, with a linear warmup for the first $10\%$ steps. The weight decay is set to $1e-2$. For all datasets, we fine-tune our model for $5$ epochs with the batch size $64$. The best model checkpoint is selected based on the performance on the development set. In Table~\ref{tab:temp_1} and Table~\ref{tab:temp_2}, we show some details of relation-specific prompts in TACRED, TACREV, ReTACRED, and SemEval.

In TACRED, reversing some relations, such as reversing ($e_s$,``organization: member\_of'', $e_o$) into ($e_o$,``organization: member'', $e_s$), can lead to better prompts. Hence, in our experiments, we apply two settings for PTR, one is to build prompts without changing any relations, and the other is to build prompts after reversing a part of relations. For models tuning with some reversed relations, we name them ``\ldots (Reversed)''.

\subsection{Comparison between PTR and Fine-Tuning Methods}

In this part, we compare PTR with several typical models for relation classification, including:

(1) \textbf{Learning models from scratch:} for relation classification, the typical approach is learning neural models from scratch. Here we choose \textsc{PA-LSTM}~\cite{zhang2017position} and \textsc{C-GCN}~\cite{zhang2018graph} as our baselines, as these two models are the most effective ones based on recurrent neural networks and graph neural networks.

(2) \textbf{Fine-tuning pre-trained models:} as PLMs have achieved promising results on various NLP tasks, some efforts are also devoted to fine-tuning PLMs for relation classification. For fine-tuning vanilla PLMs, we directly select \textsc{RoBERTa\_large} as our baseline.

(3) \textbf{Fine-tuning knowledge-enhanced pre-trained models:} Considering the entity information is quite important for models to understand relational semantics, a series of knowledge-enhanced PLMs have been further explored, which use knowledge bases as additional information to enhance PLMs. For knowledge-enhanced PLMs, we select \textsc{SpanBERT}~\cite{joshi2020spanbert}, \textsc{KnowBERT}~\cite{peters2019knowledge}, \textsc{LUKE}~\cite{yamada2020luke}, and \textsc{MTB}~\cite{baldini-soares-etal-2019-matching} as our baselines. And these four baselines are typical models that use knowledge to enhance learning objectives, input features, model architectures, and pre-training strategies respectively.

For the above baselines, we report the results of their papers as well as some results reproduced by \citet {zhou2021improved}. Table~\ref{tab:all} shows the overall performance on four datasets. From the table, we can see that:

\begin{table*}[t]
\small
\center
\scalebox{0.9}{
\begin{tabular}{p{.14\textwidth}|p{.4\textwidth}|p{.055\textwidth}|p{.055\textwidth}|p{.08\textwidth}|p{.08\textwidth}|p{.09\textwidth}}
\toprule
Model                & Input Example                                                                                                                                                                  & Extra Layer & Extra Label & TACRED & TACREV & ReTACRED    \\
\midrule
\multicolumn{7}{c}{Using randomly initialized markers for prompts}               \\
\midrule
\textsc{ENT Marker (cls)}         & {\texttt{[CLS]}}{\texttt{[E1]}} Mark Twain {\texttt{[/E1]}} was born in {\texttt{[E2]}} Florida {\texttt{[/E2]}} . {\texttt{[SEP]}}                                                                                      &     w/o        & w/o                  & 69.4 & 79.8 & 88.4 \\
\midrule
\textsc{ENT Marker}         & {\texttt{[CLS]}}{\texttt{[E1]}} Mark Twain {\texttt{[/E1]}} was born in {\texttt{[E2]}} Florida {\texttt{[/E2]}} . {\texttt{[SEP]}}                                                                                      &   w/          & w/o                  & 70.7  & 80.2 & 90.5 \\
\midrule
\textsc{TYP Marker}          & {\texttt{[CLS]}} {\texttt{[E1:Person]}} Mark Twain {\texttt{[/E1:Person]}} was born in {\texttt{[E2:State]}} Florida {\texttt{[/E2:State]}} . {\texttt{[SEP]}}                                                          &     w/         & w/                   & 71.0 & 80.8 & 90.5 \\
\midrule
\multicolumn{7}{c}{Using the vocabulary of PLMs for prompts}               \\
\midrule
\textsc{ENT Marker (punct)} & {\texttt{[CLS]}} \texttt{@} Mark Twain \texttt{@} was born in \texttt{\#} Florida \texttt{\#} . {\texttt{[SEP]}} & w/ &       w/o           & 71.4 & 81.2 & 90.5 \\

\midrule
\textsc{TYP Marker (punct)}  & {\texttt{[CLS]}} \texttt{@ * person *} Mark Twain \texttt{@} was born in \texttt{\# \textasciicircum~state \textasciicircum} Florida \texttt{\#} . {\texttt{[SEP]}}                                                                    &      w/       & w/                   & \underline{74.6} & \underline{83.2} & \underline{91.1}\\
\midrule
\textsc{AdaPrompt}  & {\texttt{[CLS]}} Mark Twain was born in Florida . Florida was  {\texttt{[MASK]}} of Mark Twain {\texttt{[SEP]}}                                                                    &      w/o       & w/o                   & - & 80.8 & -\\
\midrule
\textsc{PTR}                   & {\texttt{[CLS]}} Mark Twain  was born in  Florida . the {\texttt{[MASK]}} Mark Twain  {\texttt{[MASK]}} the {\texttt{[MASK]}}  Florida  {\texttt{[SEP]}}  &     w/o        &     w/o             & 72.4 & 81.4 & 90.9\\
\midrule
\textsc{PTR \quad\quad\quad\quad(Reversed)}        & {\texttt{[CLS]}} Mark Twain  was born in  Florida . the {\texttt{[MASK]}} Florida {\texttt{[MASK]}}  the {\texttt{[MASK]}} Mark Twain{\texttt{[SEP]}}  &      w/o &         w/o         & \textbf{75.9} & \textbf{83.9} & \textbf{91.9} \\
\bottomrule
\end{tabular}
}
\caption{$F_1$ scores (\%) of various models using prompts on the datasets TACRED, TACREV, and ReTACRED. All these models are based on \textsc{RoBERTa\_large} or \textsc{BERT\_large}. For each model, we also provide the processed input of an example sentence ``Mark Twain was born in Florida''. We also provide the information that whether these models add extra neural layers to generate features or use extra human annotations to provide more information. ``w/o'' indicates that the model does not use additional human annotations or neural layers, yet ``w/'' indicates the model requires additional human annotations or neural layers. The best results are bold, and the best results under normal settings (without reversing relations) are underlined.}
\label{tab:marker}
\end{table*}

\begin{table*}[t]
\small
\centering
\scalebox{0.9}{
\begin{tabular}{p{.14\textwidth}|cccc|cccc|cccc}
\toprule
       Model & \multicolumn{4}{c|}{TACRED} & \multicolumn{4}{c|}{TACREV} & \multicolumn{4}{c}{ReTACRED}\\
       &8&16&32&Mean&8&16&32&Mean&8&16&32&Mean \\
\midrule
\textsc{ENT Marker (punct)}    & 27.0 & 31.3 & 31.9 & 30.1 & 27.4 & 31.2 & 32.0 & 30.2 & 44.1 & 53.7 & 59.9   & 52.6   \\
\midrule
\textsc{TYP Marker (punct)}    & \textbf{28.9} & \textbf{32.0} & \textbf{32.4} & \textbf{31.1} & 27.6 & 31.2 & 32.0 & 30.3 & 44.8 & 54.1 & 60.0  & 53.0     \\
\midrule
\textsc{AdaPrompt}  & - & - & - & - & 25.2 & 27.3 & 30.8 & 27.8 & - & - & - & -   \\
\midrule
PTR    & 28.1 & 30.7 & 32.1 & 30.3 & \textbf{28.7} & \textbf{31.4} & \textbf{32.4} & \textbf{30.8} & \textbf{51.5} & \textbf{56.2} & \textbf{62.1} & \textbf{56.6}  \\
\bottomrule
\end{tabular}
}
\caption{$F_1$ scores (\%) of various models with different sizes of training instances on the datasets TACRED, TACREV, and ReTACRED.}
\label{tab:few_result}
\end{table*}

(1) Conventional methods design complicated models, including various recurrent neural layers, attention neural layers, and graph neural layers. Although learning these complicated models from scratch have achieved effective results, the vanilla PLM \textsc{RoBERTa\_large} can easily outperform them, indicating that the rich knowledge of PLMs captured from large-scale unlabeled data is important for downstream tasks. 

(2) Within PLMs, those knowledge-enhanced PLMs defeat the vanilla PLM~\textsc{RoBERTa\_large}. On the one hand, this shows that introducing extra task-specific knowledge to enhance models is effective. On the other hand, this indicates that simply fine-tuning PLMs cannot completely cover the knowledge required for downstream tasks. 

(3) Within knowledge-enhanced PLMs, \textsc{LUKE} and \textsc{KnowBERT} are better than other models. This is because these two models respectively use knowledge as data augmentation and architecture enhancement in the fine-tuning phase. In contrast, \textsc{SpanBERT} and \textsc{MTB} inject task-specific knowledge into parameters in the pre-training phase and then fine-tune their parameters without extra knowledge. All the above results indicate that even if the task-specific knowledge is already contained in PLMs, it is difficult for fine-tuning to stimulate the knowledge for downstream tasks.

(4) Either for the normal setting or the setting of reversing some relations, PTR achieves significant improvements than all baselines. Even compared with those knowledge-enhanced models, PTR still outperforms these models on TACRED. 

In general, all these experimental results show that \textbf{the pre-training phase enables PLMs to capture sufficient task-specific knowledge, how to stimulate the knowledge for downstream tasks is crucial.} Compared with typical fine-tuning methods, PTR can better stimulate task-specific knowledge.

\begin{figure*}[t]
\centering
\subfigure[K=8]{
\begin{minipage}[t]{0.32\linewidth}
\centering
\includegraphics[width=1.0\linewidth]{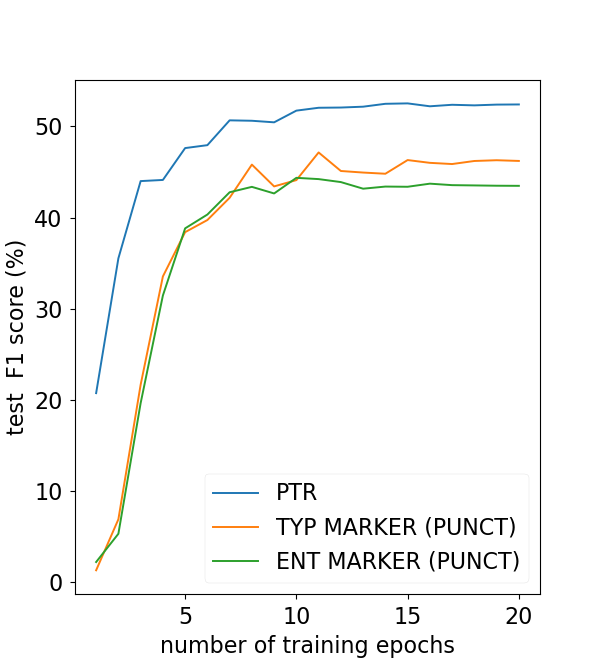}
\end{minipage}%
}%
\subfigure[K=16]{
\begin{minipage}[t]{0.32\linewidth}
\centering
\includegraphics[width=1.0\linewidth]{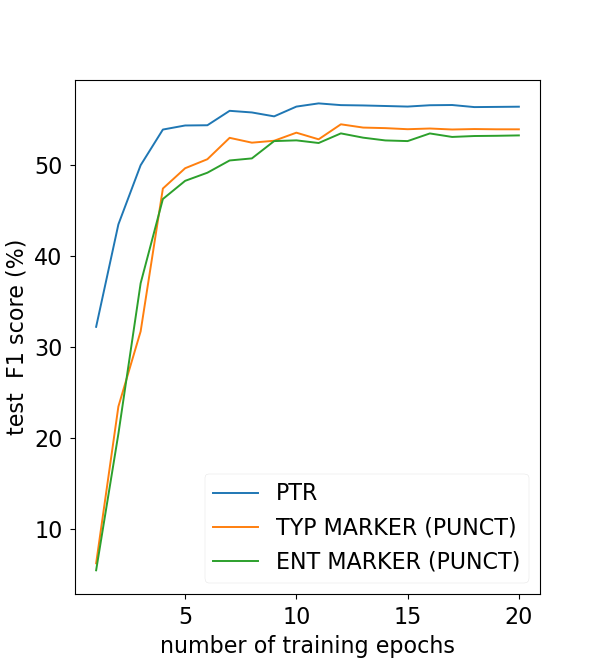}
\end{minipage}%
}%
\subfigure[K=32]{
\begin{minipage}[t]{0.32\linewidth}
\centering
\includegraphics[width=1.0\linewidth]{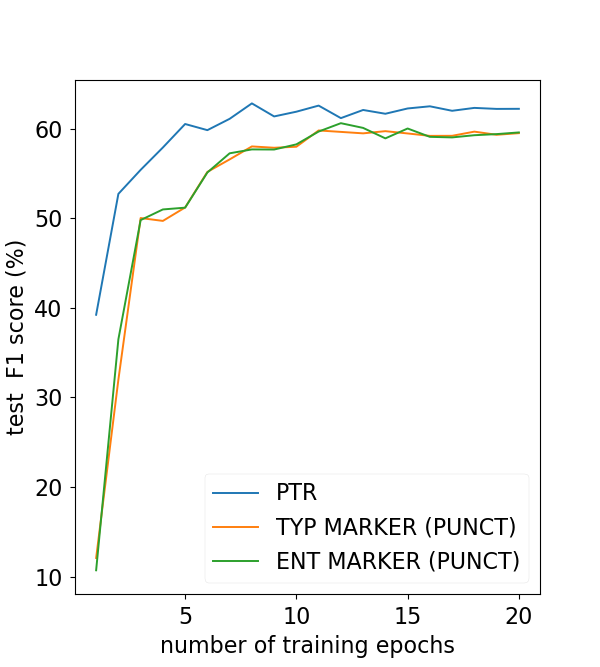}
\end{minipage}
}%
\centering
\caption{Changes in $F_1$ scores (\%) with increasing number of training epochs on the dataset ReTACRED.}
\label{fig:few_trend}
\end{figure*}

\begin{table}[t]
\center
\small
\scalebox{0.9}{
\begin{tabular}{l|c|c}
\toprule
Method        & Normal & Reversed \\
\midrule
\textsc{ENT Marker (punct)} & 71.4   & 71.6 (+ 0.2)    \\
\textsc{TYP Marker (punct)}  & 74.6   & 76.0 (+ 1.4)  \\
\textsc{PTR}           & 72.4   & 75.9 (+ \textbf{3.5})  \\
\bottomrule
\end{tabular}}
\caption{The $F_1$ scores (\%) of different models before and after reversing a part of relations in TACRED.}
\label{tab:reversed}
\end{table}

\subsection{Comparison between PTR and Prompt Tuning Methods}
\label{sec:effect_of_sub_prompts}

In this part, we compare PTR with several recent methods using prompts for relation classification, including:

(1) \textbf{Fine-tuning PLMs with entity markers:} for relation classification, some methods add additional entity markers to the input as additional information, which is similar to prompt tuning. Among these marker-based models, \textsc{ENT Marker} methods inject special symbols to index the positions of entities and \textsc{TYP Marker} methods additionally introduce the type information of entities. Intrinsically, the \textsc{ENT Marker} methods are similar to prompting by introducing extra serial information to indicate the position of special tokens, i.e., named entities. Analogously, the \textsc{TYP Marker} methods could be regarded as a type of template for prompts but require additional annotation of type information. 

``\ldots \textsc{(punct)}'' indicates using the vocabulary of PLMs to build markers. ``\ldots \textsc{(cls)}'' indicates using the hidden vector of $\texttt{[CLS]}$ for classification. Other models use additional neural layers to aggregate all entity features for classification. For more details of marker-based models, we refer to the paper of \citet{zhou2021improved}.

(2) \textbf{Prompt tuning for relation classification:} \citet{chen2021adaprompt} introduce an adaptive label words selection mechanism for prompt tuning, which scatters the relation label into a variable number of label words to handle the complex multiple label space. The model proposed by \citet{chen2021adaprompt} is named \textsc{AdaPrompt}. As \textsc{AdaPrompt} has not released its code, we only show its reported results on TACREV.

In Table~\ref{tab:marker}, we can find that using the vocabulary of PLMs for prompts can lead to better results. By designing particular sub-prompts, our PTR method has the same effect as the \textsc{TYP Marker} methods. Furthermore, PTR establishes the effectiveness in experiments without additional human annotations and neural layers. Note that in the training procedure, \textsc{TYP Marker} methods optimize the parameters of the encoder and additional output layer, PTR just optimizes the encoder and several learnable tokens in prompts.

To further compare the models using the vocabulary of PLMs for prompts, we introduce the few-shot learning scenario. Following the few-shot setting in \citet{gao2020making}, we sample $K$ training instances and $K$ validation instances per class from the original training set and development set, and evaluate models on the original test set. We set $K$ from $\{8,16,32\}$ respectively and use a fixed set of 5 random seeds to sample instances.

From Table~\ref{tab:few_result}, we find that although PTR performs poorer than \textsc{TYP Marker (punct)} using the full training dataset, PTR achieves comparable or even better results in the few-shot scenario, especially in ReTACRED where the data has fewer annotation errors. It is worth mentioning that PTR does not use extra human annotations to bring entity types, but to stimulate the type knowledge in PLMs. Compared with \textsc{TYP Marker (punct)} and $\textsc{AdaPrompt}$, \textbf{we attribute our success to the use of both human knowledge and model knowledge}. As shown in Figure~\ref{fig:few_trend}, besides effectiveness, PTR can lead to faster convergence.  TThis indicates the \textbf{potential positive effect of prompt tuning and logic rules on the convergence}.

\subsection{Effect of Reversing Relations}
We introduce the \textit{reversed} setting of TACRED, TACREV, ReTACRED in Section~\ref{sec:imp} and report the corresponding results in Table~\ref{tab:all}. Surprisingly, we find that after manipulating the labels into the reversed version, PTR gains significant improvements on all the three TACRED datasets. The results demonstrate the tremendous potential for prompt-based tuning. It suggests that small modification of prompts may result in a huge impact on whether the prior knowledge distributed in PLMs can be stimulated. Specifically, as we only change the direction of the relations and yield such improvements, prompts are position-aware.
We further apply the reversed relation setting into another two baselines, \textsc{Entity Marker} and \textsc{Typed Marker}. As shown in Table~\ref{tab:reversed} , these two methods yield 0.2\% and 1.4\% absolute improvements on TACRED, which are less than the improvements PTR gains (3.5\%) but still considerable. The results indicate that PLMs may not encode the bidirectional relation into ``symmetric'' representations. 
Based on PTR, we believe more efforts about the construction of effective prompts should be made. And theoretical analysis to better understand the mechanisms underlying prompting strategies is also urgently needed.

\section{Related Work}

Various recent PLMs like GPT~\cite{radfordimproving}, BERT~\cite{devlin2019bert}, RoBERTa~\cite{liu2019roberta} and T5~\cite{raffel2020exploring} provide a new approach to utilize large-scale unlabeled data for NLP tasks. Although these PLMs can capture rich knowledge~\cite{jawahar2019does,hewitt2019structural,petroni2019language,yenicelik2020does} from massive corpora, a fine-tuning process with extra task-specific data is still required to transfer their knowledge for downstream tasks. From dialogue~\cite{zhang2019dialogpt}, summarization~\cite{zhang2019pegasus,liu-lapata-2019-text}, question answering~\cite{adiwardana2020humanlike}, to text classification~\cite{baldini-soares-etal-2019-matching,peng2020learning, ding2021prototypical}, fine-tuned PLMs have been demonstrated their effectiveness on almost all important NLP tasks. Besides fine-tuning language models for specific tasks, recent studies have explored better optimization and regularization techniques to improve fine-tuning~\cite{lee2019mixout,dodge2020fine}. 

Despite the success of fine-tuning PLMs, there is a big gap between pre-training objectives and fine-tuning objectives. In the pre-training phase, sequential language models and masked language models are used to learn PLMs. In the fine-tuning phase, the optimization objectives are task-specific and different tasks may have quite different objective forms. In GPT-3~\cite{brown2020language}, prompt tuning has been proposed and drawn much attention. By leveraging language prompts as contexts, downstream tasks can be expressed as some objectives similar to pre-training objectives. Through a series of research work on knowledge probing~\cite{trinh2018simple,petroni2019language,davison2019commonsense}, language prompts have been proven to effectively stimulate knowledge from PLMs. Moreover, human-picked prompts have achieved promising results on few-class classification tasks such as sentiment classification and natural language inference~\cite{schick2020exploiting,liu2021gpt}. 

To avoid labor-intensive prompt design, automatic prompt search has been extensively explored. As a typical prompt consists of two parts: a template and a set of label words, \citet{schick2020automatically,schick2020exploiting} first explore automatic identification of label words for human-picked templates. \citet{shin2020eliciting} further explore gradient-guided search to automatically generate both templates and label words. \citet{gao2020making} take seq-to-seq models to generate prompt candidates, and then use each of them for prompt tuning and verify their effectiveness on development sets. Compared with human-picked prompts, most auto-generated prompts cannot achieve comparable performance. Recently, some continuous prompts have also been proposed~\cite{li2021prefix,lester2021power}, which directly use a series of learnable continuous embeddings as prompt templates and get rid of the trouble of designing prompts. These completely continuous prompts work well on those large-scale PLMs with billions of parameters, yet cannot stably work on normal PLMs.

In this paper, we propose PTR to conduct prompt tuning with rules. Based on several human-picked sub-prompts, PTR can use predefined logic rules to compose sub-prompts into complete task-specific prompts. As compared with the above-mentioned prompt tuning methods, PTR achieves a good balance among model efficiency, model effectiveness, model generalization, and human workload.

\section{Conclusion}
In this paper, we propose prompt tuning with rules (PTR) for many-class text classification. By composing sub-prompts into task-specific prompts according to logic rules, prior human knowledge can be encoded into prompt tuning. Meanwhile, the introduction of sub-prompts can further alleviate the difficulty of designing templates and sets of label words. The experimental results on relation classification, a typical many-class classification task, show that PTR can significantly outperform existing state-of-the-art baselines without introducing any additional model layers, manual annotations, and augmented data. In the future, we will further combine the existing exploration of auto-generated prompts and our PTR, which will lead to a more practical approach to prompt tuning.

\bibliography{custom}

\begin{thebibliography}{40}
\expandafter\ifx\csname natexlab\endcsname\relax\def\natexlab#1{#1}\fi

\bibitem[{Adiwardana et~al.(2020)Adiwardana, Luong, So, Hall, Fiedel,
  Thoppilan, Yang, Kulshreshtha, Nemade, Lu et~al.}]{adiwardana2020humanlike}
Daniel Adiwardana, Minh-Thang Luong, David~R So, Jamie Hall, Noah Fiedel, Romal
  Thoppilan, Zi~Yang, Apoorv Kulshreshtha, Gaurav Nemade, Yifeng Lu, et~al.
  2020.
\newblock \href {https://arxiv.org/abs/2001.09977} {Towards a human-like
  open-domain chatbot}.
\newblock \emph{arXiv preprint arXiv:2001.09977}.

\bibitem[{Alt et~al.(2020)Alt, Gabryszak, and Hennig}]{alt2020tacred}
Christoph Alt, Aleksandra Gabryszak, and Leonhard Hennig. 2020.
\newblock \href {https://www.aclweb.org/anthology/2020.acl-main.142} {Tacred
  revisited: A thorough evaluation of the tacred relation extraction task}.
\newblock In \emph{Proceedings of ACL}, pages 1558--1569.

\bibitem[{Baldini~Soares et~al.(2019)Baldini~Soares, FitzGerald, Ling, and
  Kwiatkowski}]{baldini-soares-etal-2019-matching}
Livio Baldini~Soares, Nicholas FitzGerald, Jeffrey Ling, and Tom Kwiatkowski.
  2019.
\newblock \href {https://www.aclweb.org/anthology/P19-1279} {Matching the
  blanks: Distributional similarity for relation learning}.
\newblock In \emph{Proceedings of ACL}, pages 2895--2905.

\bibitem[{Brown et~al.(2020)Brown, Mann, Ryder, Subbiah, Kaplan, Dhariwal,
  Neelakantan, Shyam, Sastry, Askell et~al.}]{brown2020language}
Tom~B Brown, Benjamin Mann, Nick Ryder, Melanie Subbiah, Jared Kaplan, Prafulla
  Dhariwal, Arvind Neelakantan, Pranav Shyam, Girish Sastry, Amanda Askell,
  et~al. 2020.
\newblock \href
  {https://proceedings.neurips.cc/paper/2020/file/1457c0d6bfcb4967418bfb8ac142f64a-Paper.pdf}
  {Language models are few-shot learners}.
\newblock In \emph{Proceedings of NIPS}, pages 1877--1901.

\bibitem[{Chen et~al.(2021)Chen, Xie, Zhang, Yan, Deng, Tan, Huang, Si, and
  Chen}]{chen2021adaprompt}
Xiang Chen, Xin Xie, Ningyu Zhang, Jiahuan Yan, Shumin Deng, Chuanqi Tan, Fei
  Huang, Luo Si, and Huajun Chen. 2021.
\newblock \href {https://arxiv.org/abs/2104.07650} {Adaprompt: Adaptive
  prompt-based finetuning for relation extraction}.
\newblock \emph{arXiv preprint arXiv:2104.07650}.

\bibitem[{Davison et~al.(2019)Davison, Feldman, and
  Rush}]{davison2019commonsense}
Joe Davison, Joshua Feldman, and Alexander~M Rush. 2019.
\newblock \href {https://arxiv.org/abs/1909.00505} {Commonsense knowledge
  mining from pretrained models}.
\newblock In \emph{Proceedings of EMNLP-IJCNLP}, pages 1173--1178.

\bibitem[{Devlin et~al.(2019)Devlin, Chang, Lee, and
  Toutanova}]{devlin2019bert}
Jacob Devlin, Ming-Wei Chang, Kenton Lee, and Kristina Toutanova. 2019.
\newblock \href {https://www.aclweb.org/anthology/N19-1423.pdf} {Bert:
  Pre-training of deep bidirectional transformers for language understanding}.
\newblock In \emph{Proceedings of NAACL-HLT}, pages 4171--4186.

\bibitem[{Ding et~al.(2021)Ding, Wang, Fu, Xu, Wang, Xie, Shen, Huang, Zheng,
  and Zhang}]{ding2021prototypical}
Ning Ding, Xiaobin Wang, Yao Fu, Guangwei Xu, Rui Wang, Pengjun Xie, Ying Shen,
  Fei Huang, Hai-Tao Zheng, and Rui Zhang. 2021.
\newblock \href {https://openreview.net/forum?id=aCgLmfhIy_f} {Prototypical
  representation learning for relation extraction}.
\newblock In \emph{Proceedings of ICLR}.

\bibitem[{Dodge et~al.(2020)Dodge, Ilharco, Schwartz, Farhadi, Hajishirzi, and
  Smith}]{dodge2020fine}
Jesse Dodge, Gabriel Ilharco, Roy Schwartz, Ali Farhadi, Hannaneh Hajishirzi,
  and Noah Smith. 2020.
\newblock \href {https://arxiv.org/abs/2002.06305} {Fine-tuning pretrained
  language models: Weight initializations, data orders, and early stopping}.
\newblock \emph{arXiv preprint arXiv:2002.06305}.

\bibitem[{Gao et~al.(2020)Gao, Fisch, and Chen}]{gao2020making}
Tianyu Gao, Adam Fisch, and Danqi Chen. 2020.
\newblock \href {https://arxiv.org/pdf/2012.15723} {Making pre-trained language
  models better few-shot learners}.
\newblock \emph{arXiv preprint arXiv:2012.15723}.

\bibitem[{Han et~al.(2019)Han, Gao, Yao, Ye, Liu, and
  Sun}]{han-etal-2019-opennre}
Xu~Han, Tianyu Gao, Yuan Yao, Deming Ye, Zhiyuan Liu, and Maosong Sun. 2019.
\newblock \href {https://doi.org/10.18653/v1/D19-3029} {{O}pen{NRE}: An open
  and extensible toolkit for neural relation extraction}.
\newblock In \emph{Proceedings of EMNLP-IJCNLP}, pages 169--174.

\bibitem[{Hendrickx et~al.(2010)Hendrickx, Kim, Kozareva, Nakov,
  {\'O}~S{\'e}aghdha, Pad{\'o}, Pennacchiotti, Romano, and
  Szpakowicz}]{hendrickx2010semeval}
Iris Hendrickx, Su~Nam Kim, Zornitsa Kozareva, Preslav Nakov, Diarmuid
  {\'O}~S{\'e}aghdha, Sebastian Pad{\'o}, Marco Pennacchiotti, Lorenza Romano,
  and Stan Szpakowicz. 2010.
\newblock \href {https://www.aclweb.org/anthology/S10-1006/} {{S}em{E}val-2010
  task 8: Multi-way classification of semantic relations between pairs of
  nominals}.
\newblock In \emph{Proceedings of SemEval}, pages 33--38.

\bibitem[{Hewitt and Manning(2019)}]{hewitt2019structural}
John Hewitt and Christopher~D Manning. 2019.
\newblock \href {https://www.aclweb.org/anthology/N19-1419"} {A structural
  probe for finding syntax in word representations}.
\newblock In \emph{Proceedings of NAACL}, pages 4129--4138.

\bibitem[{Jawahar et~al.(2019)Jawahar, Sagot, and Seddah}]{jawahar2019does}
Ganesh Jawahar, Beno{\^\i}t Sagot, and Djam{\'e} Seddah. 2019.
\newblock \href {https://www.aclweb.org/anthology/P19-1356} {What does bert
  learn about the structure of language?}
\newblock In \emph{Proceedings of ACL}, pages 3651--3657.

\bibitem[{Joshi et~al.(2020)Joshi, Chen, Liu, Weld, Zettlemoyer, and
  Levy}]{joshi2020spanbert}
Mandar Joshi, Danqi Chen, Yinhan Liu, Daniel~S Weld, Luke Zettlemoyer, and Omer
  Levy. 2020.
\newblock \href {https://www.aclweb.org/anthology/2020.tacl-1.5} {Spanbert:
  Improving pre-training by representing and predicting spans}.
\newblock \emph{TACL}, 8:64--77.

\bibitem[{Lee et~al.(2019)Lee, Cho, and Kang}]{lee2019mixout}
Cheolhyoung Lee, Kyunghyun Cho, and Wanmo Kang. 2019.
\newblock \href {https://openreview.net/forum?id=HkgaETNtDB} {Mixout: Effective
  regularization to finetune large-scale pretrained language models}.
\newblock In \emph{Proceedings of ICLR}.

\bibitem[{Lester et~al.(2021)Lester, Al-Rfou, and Constant}]{lester2021power}
Brian Lester, Rami Al-Rfou, and Noah Constant. 2021.
\newblock \href {https://arxiv.org/abs/2104.08691} {The power of scale for
  parameter-efficient prompt tuning}.
\newblock \emph{arXiv preprint arXiv:2104.08691}.

\bibitem[{Li and Liang(2021)}]{li2021prefix}
Xiang~Lisa Li and Percy Liang. 2021.
\newblock \href {https://arxiv.org/abs/2101.00190} {Prefix-tuning: Optimizing
  continuous prompts for generation}.
\newblock \emph{arXiv preprint arXiv:2101.00190}.

\bibitem[{Liu et~al.(2021)Liu, Zheng, Du, Ding, Qian, Yang, and
  Tang}]{liu2021gpt}
Xiao Liu, Yanan Zheng, Zhengxiao Du, Ming Ding, Yujie Qian, Zhilin Yang, and
  Jie Tang. 2021.
\newblock \href {https://arxiv.org/abs/2103.10385} {Gpt understands, too}.
\newblock \emph{arXiv preprint arXiv:2103.10385}.

\bibitem[{Liu and Lapata(2019)}]{liu-lapata-2019-text}
Yang Liu and Mirella Lapata. 2019.
\newblock \href {https://www.aclweb.org/anthology/D19-1387/} {Text
  summarization with pretrained encoders}.
\newblock In \emph{Proceedings of EMNLP}, pages 3730--3740.

\bibitem[{Liu et~al.(2019)Liu, Ott, Goyal, Du, Joshi, Chen, Levy, Lewis,
  Zettlemoyer, and Stoyanov}]{liu2019roberta}
Yinhan Liu, Myle Ott, Naman Goyal, Jingfei Du, Mandar Joshi, Danqi Chen, Omer
  Levy, Mike Lewis, Luke Zettlemoyer, and Veselin Stoyanov. 2019.
\newblock \href {https://arxiv.org/abs/1907.11692} {Roberta: A robustly
  optimized bert pretraining approach}.
\newblock \emph{arXiv preprint arXiv:1907.11692}.

\bibitem[{Peng et~al.(2020)Peng, Gao, Han, Lin, Li, Liu, Sun, and
  Zhou}]{peng2020learning}
Hao Peng, Tianyu Gao, Xu~Han, Yankai Lin, Peng Li, Zhiyuan Liu, Maosong Sun,
  and Jie Zhou. 2020.
\newblock \href {https://doi.org/10.18653/v1/2020.emnlp-main.298} {{L}earning
  from {C}ontext or {N}ames? {A}n {E}mpirical {S}tudy on {N}eural {R}elation
  {E}xtraction}.
\newblock In \emph{Proceedings of EMNLP}, pages 3661--3672.

\bibitem[{Peters et~al.(2019)Peters, Neumann, Logan, Schwartz, Joshi, Singh,
  and Smith}]{peters2019knowledge}
Matthew~E Peters, Mark Neumann, Robert Logan, Roy Schwartz, Vidur Joshi, Sameer
  Singh, and Noah~A Smith. 2019.
\newblock \href {https://www.aclweb.org/anthology/D19-1005} {Knowledge enhanced
  contextual word representations}.
\newblock In \emph{Proceedings of EMNLP-IJCNLP}, pages 43--54.

\bibitem[{Petroni et~al.(2019)Petroni, Rockt{\"a}schel, Riedel, Lewis, Bakhtin,
  Wu, and Miller}]{petroni2019language}
Fabio Petroni, Tim Rockt{\"a}schel, Sebastian Riedel, Patrick Lewis, Anton
  Bakhtin, Yuxiang Wu, and Alexander Miller. 2019.
\newblock \href {https://www.aclweb.org/anthology/D19-1250.pdf} {Language
  models as knowledge bases?}
\newblock In \emph{Proceedings of EMNLP}, pages 2463--2473.

\bibitem[{Qiu et~al.(2020)Qiu, Sun, Xu, Shao, Dai, and Huang}]{qiu2020pre}
Xipeng Qiu, Tianxiang Sun, Yige Xu, Yunfan Shao, Ning Dai, and Xuanjing Huang.
  2020.
\newblock \href {https://arxiv.org/abs/2003.08271} {Pre-trained models for
  natural language processing: A survey}.
\newblock \emph{Science China Technological Sciences}, pages 1--26.

\bibitem[{Radford et~al.(2018)Radford, Narasimhan, Salimans, and
  Sutskever}]{radfordimproving}
Alec Radford, Karthik Narasimhan, Tim Salimans, and Ilya Sutskever. 2018.
\newblock \href
  {https://www.cs.ubc.ca/~amuham01/LING530/papers/radford2018improving.pdf}
  {Improving language understanding by generative pre-training}.
\newblock \emph{OpenAI}.

\bibitem[{Raffel et~al.(2020)Raffel, Shazeer, Roberts, Lee, Narang, Matena,
  Zhou, Li, and Liu}]{raffel2020exploring}
Colin Raffel, Noam Shazeer, Adam Roberts, Katherine Lee, Sharan Narang, Michael
  Matena, Yanqi Zhou, Wei Li, and Peter~J Liu. 2020.
\newblock \href {https://www.jmlr.org/papers/volume21/20-074/20-074.pdf}
  {Exploring the limits of transfer learning with a unified text-to-text
  transformer}.
\newblock \emph{JMLR}, 21:1--67.

\bibitem[{Schick et~al.(2020)Schick, Schmid, and
  Sch{\"u}tze}]{schick2020automatically}
Timo Schick, Helmut Schmid, and Hinrich Sch{\"u}tze. 2020.
\newblock \href {https://doi.org/10.18653/v1/2020.coling-main.488}
  {Automatically identifying words that can serve as labels for few-shot text
  classification}.
\newblock In \emph{Proceedings of COLING}, pages 5569--5578.

\bibitem[{Schick and Sch{\"u}tze(2021)}]{schick2020exploiting}
Timo Schick and Hinrich Sch{\"u}tze. 2021.
\newblock \href {https://www.aclweb.org/anthology/2021.eacl-main.20}
  {Exploiting cloze-questions for few-shot text classification and natural
  language inference}.
\newblock In \emph{Proceedings of EACL}, pages 255--269.

\bibitem[{Shin et~al.(2020)Shin, Razeghi, Logan~IV, Wallace, and
  Singh}]{shin2020eliciting}
Taylor Shin, Yasaman Razeghi, Robert~L Logan~IV, Eric Wallace, and Sameer
  Singh. 2020.
\newblock \href {https://www.aclweb.org/anthology/2020.emnlp-main.346.pdf}
  {Eliciting knowledge from language models using automatically generated
  prompts}.
\newblock In \emph{Proceedings of EMNLP}, pages 4222--4235.

\bibitem[{Stoica et~al.(2021)Stoica, Platanios, and P{\'o}czos}]{stoica2021re}
George Stoica, Emmanouil~Antonios Platanios, and Barnab{\'a}s P{\'o}czos. 2021.
\newblock \href {https://arxiv.org/abs/2104.08398} {Re-tacred: Addressing
  shortcomings of the tacred dataset}.
\newblock \emph{arXiv preprint arXiv:2104.08398}.

\bibitem[{Trinh and Le(2018)}]{trinh2018simple}
Trieu~H Trinh and Quoc~V Le. 2018.
\newblock \href {https://arxiv.org/abs/1806.02847} {A simple method for
  commonsense reasoning}.
\newblock \emph{arXiv preprint arXiv:1806.02847}.

\bibitem[{Wolf et~al.(2020)Wolf, Chaumond, Debut, Sanh, Delangue, Moi, Cistac,
  Funtowicz, Davison, Shleifer et~al.}]{wolf2020transformers}
Thomas Wolf, Julien Chaumond, Lysandre Debut, Victor Sanh, Clement Delangue,
  Anthony Moi, Pierric Cistac, Morgan Funtowicz, Joe Davison, Sam Shleifer,
  et~al. 2020.
\newblock \href {https://www.aclweb.org/anthology/2020.emnlp-demos.6.pdf}
  {Transformers: State-of-the-art natural language processing}.
\newblock In \emph{Proceedings of EMNLP}, pages 38--45.

\bibitem[{Yamada et~al.(2020)Yamada, Asai, Shindo, Takeda, and
  Matsumoto}]{yamada2020luke}
Ikuya Yamada, Akari Asai, Hiroyuki Shindo, Hideaki Takeda, and Yuji Matsumoto.
  2020.
\newblock \href {https://www.aclweb.org/anthology/2020.emnlp-main.523} {Luke:
  Deep contextualized entity representations with entity-aware self-attention}.
\newblock In \emph{Proceedings of EMNLP}, pages 6442--6454.

\bibitem[{Yenicelik et~al.(2020)Yenicelik, Schmidt, and
  Kilcher}]{yenicelik2020does}
David Yenicelik, Florian Schmidt, and Yannic Kilcher. 2020.
\newblock \href {https://www.aclweb.org/anthology/2020.blackboxnlp-1.15} {How
  does bert capture semantics? a closer look at polysemous words}.
\newblock In \emph{Proceedings of BlackboxNLP}, pages 156--162.

\bibitem[{Zhang et~al.(2019)Zhang, Zhao, Saleh, and Liu}]{zhang2019pegasus}
Jingqing Zhang, Yao Zhao, Mohammad Saleh, and Peter~J. Liu. 2019.
\newblock \href {https://arxiv.org/abs/1912.08777} {Pegasus: Pre-training with
  extracted gap-sentences for abstractive summarization}.
\newblock In \emph{Proceedings of ICML}, pages 11328--11339.

\bibitem[{Zhang et~al.(2020)Zhang, Sun, Galley, Chen, Brockett, Gao, Gao, Liu,
  and Dolan}]{zhang2019dialogpt}
Yizhe Zhang, Siqi Sun, Michel Galley, Yen-Chun Chen, Chris Brockett, Xiang Gao,
  Jianfeng Gao, Jingjing Liu, and Bill Dolan. 2020.
\newblock \href {https://arxiv.org/abs/1911.00536} {Dialogpt: Large-scale
  generative pre-training for conversational response generation}.
\newblock In \emph{Proceedings of ACL}, pages 270--278.

\bibitem[{Zhang et~al.(2018)Zhang, Qi, and Manning}]{zhang2018graph}
Yuhao Zhang, Peng Qi, and Christopher~D. Manning. 2018.
\newblock \href {http://aclweb.org/anthology/D18-1244} {Graph convolution over
  pruned dependency trees improves relation extraction}.
\newblock In \emph{Proceedings of EMNLP}, pages 2205--2215.

\bibitem[{Zhang et~al.(2017)Zhang, Zhong, Chen, Angeli, and
  Manning}]{zhang2017position}
Yuhao Zhang, Victor Zhong, Danqi Chen, Gabor Angeli, and Christopher~D Manning.
  2017.
\newblock \href {https://nlp.stanford.edu/pubs/zhang2017tacred.pdf}
  {Position-aware attention and supervised data improve slot filling}.
\newblock In \emph{Proceedings of EMNLP}, pages 35--45.

\bibitem[{Zhou and Chen(2021)}]{zhou2021improved}
Wenxuan Zhou and Muhao Chen. 2021.
\newblock \href {https://arxiv.org/abs/2102.01373} {An improved baseline for
  sentence-level relation extraction}.
\newblock \emph{arXiv preprint arXiv:2102.01373}.

\end{thebibliography}
\bibliographystyle{acl_natbib}

\end{document}